\documentclass[journal]{IEEEtran}  

\usepackage{amsmath}
\usepackage{amssymb}
\usepackage[colorinlistoftodos]{todonotes}
\usepackage{hyperref}
\usepackage{multirow}
\usepackage{booktabs}
\usepackage[colorinlistoftodos]{todonotes}
\usepackage{siunitx}
\usepackage{url}
\usepackage{import,bm}
\usepackage[super]{nth}
\usepackage{romannum}
\usepackage{accents}
\usepackage{verbatim}
\usepackage{tikz}
\usetikzlibrary{automata, positioning, arrows, shapes.geometric, arrows.meta, positioning, calc, intersections, spath3}
\usepackage{tkz-euclide}
\usepackage{subcaption}
\usepackage{graphicx}
\usepackage{cite}
\usepackage{makecell}
\usepackage{caption}
\usepackage{ragged2e}
\usepackage{mathtools}
\usepackage{tikz}
\usetikzlibrary{automata, positioning, arrows, shapes.geometric, arrows.meta, positioning, calc, intersections, spath3}
\usepackage{tkz-euclide}
\usepackage{pifont}
\usepackage{lipsum}

\newcommand{\claim}[1]{\noindent\textbf{#1}}

\title{Primitive-Informed Sampling-Based MPC for Multi-Fingered Dexterous Manipulation}
\author{Emek Barış Küçüktabak$^{1}$, Karankumar Patel$^{1, \dagger}$, Jinda Cui$^{1}$, Zhaodong Yang$^{1,2}$, Kazuhiro Sasabuchi$^{1}$, Jun~Takamatsu$^{1}$

\thanks {$^1$ Honda Research Institute USA, San Jose, CA, USA}
\thanks {$^2$ Georgia Institute of Technology, Atlanta, GA, USA (Work done during an internship at HRI)}
\thanks {$^\dagger$ Current address: Skild AI, San Mateo, CA, USA (Work done while at HRI)}
}

\begin{document}
\bstctlcite{IEEEexample:BSTcontrol} 
\maketitle
\thispagestyle{empty}
\pagestyle{empty}

\begin{abstract}

We present a primitive-informed sampling-based model predictive control (MPC) framework for multi-fingered dexterous manipulation. Sampling-based MPC evaluates candidate control trajectories through forward simulation without requiring gradients through complex contact dynamics. However, directly sampling these trajectories in the high-dimensional joint space of a dexterous hand is inefficient and makes performance strongly dependent on the sampling distribution. Our framework biases sampling using low-dimensional manipulation primitives that encode coordinated finger motions, while simultaneously optimizing joint-level residuals to adapt these motions to the current hand–object configuration. Task-related rollout constraints reject infeasible trajectories during forward simulation, improving the effective use of the sampling budget. We evaluate the approach on a 16 DoF Allegro hand using a synchronized MuJoCo digital twin. Ablations show that both the primitive and residual are necessary for reliable continuous in-hand rotation, that increasing the sampling budget alone does not recover this coordination, and that rollout constraints substantially improve success rate. A primitive extracted for one object size transfers to other sizes and remains effective under model mismatch. The framework further supports grasping, object reorientation, and coordinated arm–hand manipulation, using primitives extracted from both a simulation-trained policy and human hand-motion data.\\Project webpage: \href{https://primitive-informed-mpc.github.io/}{primitive-informed-mpc.github.io}


\end{abstract}
\section{Introduction}
\label{sec:Intro}

Multi-fingered dexterous manipulation enables robots to reorient and reposition objects within the hand, but doing so requires coordinated control of many coupled degrees of freedom under complex contact dynamics. Small variations in joint motion can lead to qualitatively different contact outcomes, making the coordinated motion sequences required for successful manipulation difficult to discover.


Learning-based methods have demonstrated impressive dexterous behaviors, including in-hand reorientation, continuous object rotation, and transfer of policies trained in large-scale simulation~\cite{openai2018learning,akkaya2019solving,handa2023dextreme,chen2023visual,yin2023rotating}. However, these methods commonly require substantial offline training, large simulation datasets, and careful sim-to-real transfer, while adaptation to new task objectives, objects, or contact conditions may require additional training. These limitations motivate online control methods that exploit a dynamics model and optimize actions directly from the current measured state.

Model predictive control (MPC) provides an alternative by repeatedly optimizing actions from the current state using a predictive model. However, applying conventional gradient-based optimization to contact-rich manipulation is challenging because contact formation, separation, impact, and frictional transitions introduce nonsmooth and discontinuous changes in the system dynamics. Although smooth contact models and contact-implicit formulations can enable gradient-based optimization, they rely on approximations of the contact dynamics, and reliable gradient information remains difficult to obtain through contact transitions~\cite{jiang2024contact, kurtz2026inverse}.


Sampling-based MPC instead evaluates candidate action sequences through forward simulation without requiring gradients through the contact dynamics. At each control update, candidate sequences are rolled out using a dynamics model, the leading portion of the optimized sequence is applied, and optimization is repeated from the latest state estimate~\cite{williams2018mppi,howell2022mjpc,pezzato2025sampling}. This receding-horizon process provides feedback against disturbances and modeling errors while allowing task objectives and constraints to be modified without retraining.

Recent studies have demonstrated the potential of sampling-based MPC for multi-fingered dexterous manipulation. Li et al.~\cite{li2024drop} used online sampling-based planning and real-time state estimation to perform in-hand cube reorientation. Hess et al.~\cite{hess2024sampling} demonstrated physical ball rolling, flipping, and catching with a tendon-driven hand using sampling-based MPC. These results show that sufficiently informative contact sequences can emerge from online forward simulation. In both approaches, candidate hand motions are explored through direct joint-level sampling; neither incorporates a behavior-derived coordination prior into the sampling distribution. At the same time, these approaches highlight a central limitation of sampling-based MPC: the quality of the resulting behavior depends strongly on the number and distribution of sampled trajectories.

\begin{figure*}[t]
\centering
\includegraphics[width=2\columnwidth]{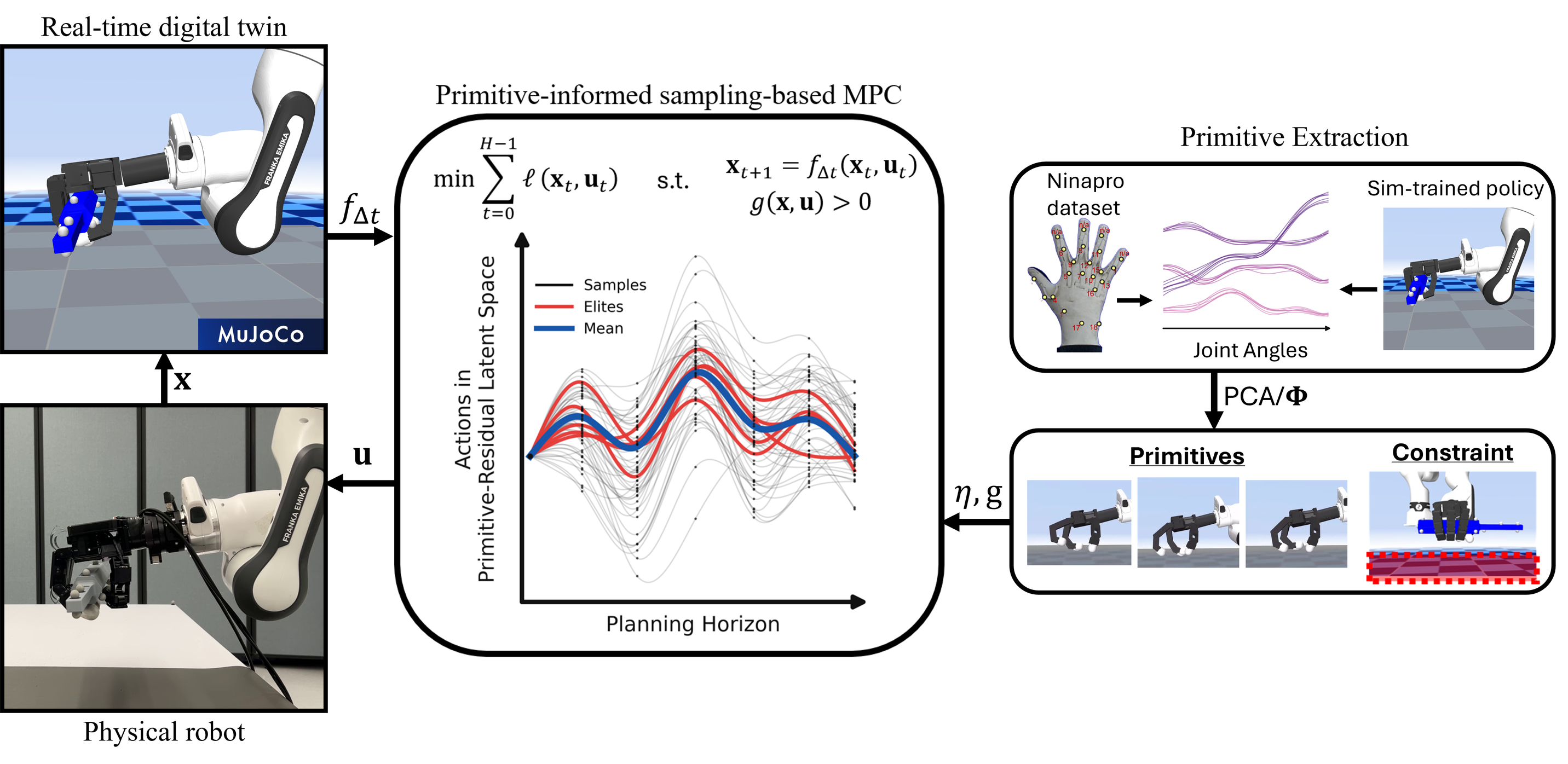}
\caption{System overview. Low-dimensional manipulation primitives bias trajectory sampling toward coordinated motions, while joint-level residuals adapt the sampled actions to the current hand–object configuration. Candidate trajectories are evaluated through forward simulation, and the optimized mean action sequence is executed on the physical system.}
\label{fig:overview}
\end{figure*}


This limitation becomes severe for multi-fingered hands. Even with spline-based trajectories, directly sampling every joint produces a high-dimensional search space in which most sampled motions lack the coordination needed to maintain a grasp or transition between contact configurations. Increasing the rollout budget can improve the chance of finding useful trajectories, but increases replanning time, reduces the controller update rate, and limits responsiveness to disturbances and model mismatch.

Learned trajectory sampling distributions have been proposed to improve the efficiency of sampling-based MPC~\cite{sacks2023learning,power2024learning}. These methods bias sampling toward low-cost trajectories and have demonstrated improved performance on navigation and collision-free reaching tasks. However, their manipulation evaluations do not consider multi-fingered hands or contact-rich object manipulation. Recent work on whole-body loco-manipulation instead reduces the effective MPC search space by planning over commands to a pretrained whole-body policy~\cite{zhang2026sumo}, but it considers manipulation with a gripper and assumes such a policy with a compact command interface is available, which may not hold for multi-fingered hands. Prior work has also shown that structured sampling can facilitate exploration of difficult dexterous behaviors~\cite{khandate2023sampling}; however, that work uses sampling-based planning to generate training distributions for reinforcement learning rather than as the deployed receding-horizon controller.

Low-dimensional structure in hand motion has long been exploited outside of MPC. Human grasping postures are well described by a small number of postural synergies~\cite{santello1998postural}, and eigengrasp representations have been used to reduce the dimensionality of grasp planning and hand control~\cite{ciocarlie2009hand}. However, these approaches have primarily been used for static grasp synthesis, and constraining motion to the synergy subspace alone limits the ability to adapt to contact variation. To our knowledge, synergy-like coordination bases have not been used as a sampling prior for online receding-horizon control on a physical hand.

In this work, we introduce a primitive-informed sampling-based MPC framework for multi-fingered dexterous manipulation, as shown in Fig.~\ref{fig:overview}. Here, a manipulation primitive refers to a low-dimensional, PCA-based coordination basis extracted from joint trajectories exhibiting the desired behavior, analogous to postural synergies.
The controller samples coefficients of this basis to generate coordinated motions and supplements them with a full-dimensional joint-level residual. The residual enables detailed joint-level exploration and adaptation to the current contact configuration and object variation. During rollout evaluation, task-related constraints reject infeasible trajectories, preventing them from influencing the sampling update and improving the effective use of the computation budget.

We evaluate the framework on a physical Allegro hand using a synchronized MuJoCo~\cite{todorov2012mujoco} digital twin. The results show that primitive guidance improves the effectiveness of the sampling budget, and that simply increasing the number of samples does not recover coordinated behavior, while residual optimization and constraint-aware evaluation enable adaptation and reliable task execution.

In summary, the contributions of this work are:
\begin{itemize}
 
\item We introduce a primitive-informed sampling-based MPC framework that jointly optimizes low-dimensional primitive coefficients and joint-level residuals, while rejecting infeasible rollouts through task-related constraints.
 
\item We conduct systematic hardware experiments that isolate the effects of the manipulation primitive, joint-level residual, rollout constraints, and rollout budget.
 

\item We show that a rotation primitive extracted using one object size remains effective for two additional sizes and under $\pm 50\%$ model-parameter perturbations, and demonstrate the framework on grasping, object reorientation, and coordinated arm-hand manipulation involving reaching, grasping, and transport.
 
\end{itemize}

\section{Methods}
\label{sec:methods}

\subsection{Problem Formulation and Sampling-Based MPC}
\label{sec:mpc}

The planner predicts the evolution of the robot-object system by rolling out candidate control trajectories in MuJoCo. The dynamics over one model timestep are written as
\begin{equation}
    \mathbf{x}_{t+1}
    =
    f_{\Delta t}(\mathbf{x}_t,\mathbf{u}_t),
    \label{eq:dynamics}
\end{equation}
where $\mathbf{x}_t\in\mathbb{R}^{n_x}$ is the robot-object state at time $t$, $\mathbf{u}_t\in\mathbb{R}^{n_u}$ is the commanded action, and $f_{\Delta t}$ denotes the MuJoCo transition over timestep $\Delta t$.

Starting from the current state, the planner searches for a control trajectory that minimizes a finite-horizon objective,
\begin{equation}
    J(\mathbf{U};\mathbf{x}_0)
    =
    \sum_{t=0}^{H-1}
    \ell(\mathbf{x}_t,\mathbf{u}_t),
    \label{eq:objective}
\end{equation}
where $\mathbf{U}$ denotes the candidate control trajectory, $H$ is the planning horizon, and $\ell(\cdot)$ is the running cost. The costs are constructed from task-relevant quantities, such as the error between the desired and predicted object poses.

Rather than optimizing an independent action at every timestep, the planner represents the control trajectory using $K$ latent spline knots~\cite{howell2022mjpc},
\begin{equation}
    \boldsymbol{\eta}(t)
    =
    \mathcal{S}\!\left(
        t;\boldsymbol{\eta}_1,\ldots,\boldsymbol{\eta}_K
    \right),
    \qquad
    \mathbf{u}_t = g\!\left(\boldsymbol{\eta}(t)\right),
    \label{eq:latent_spline}
\end{equation}
where $\boldsymbol{\eta}_k$ is the latent command at knot $k$, $\mathcal{S}(\cdot)$ interpolates between the knots, and $g(\cdot)$ maps the interpolated latent command to the full robot action. 
Because $K$ is substantially smaller than the number of timesteps in the planning horizon, this representation reduces the number of optimization variables and produces temporally smooth candidate motions.

We optimize the spline parameters using the cross-entropy method (CEM)~\cite{li2024drop, pinneri2020sample}. At each planning update, the planner samples $N$ candidate sequences from a distribution centered around the current nominal trajectory. Each candidate is reconstructed into a control trajectory, rolled out in MuJoCo, and evaluated using~\eqref{eq:objective}. The $N_e$ lowest-cost candidates form the elite set, whose statistics are used to update the mean $\boldsymbol{\mu}$ and standard deviation $\boldsymbol{\sigma}$ of the sampling distribution. The standard deviation determines the sampling spread around the current mean for each latent action dimension. 
The optimized trajectory from the previous update is shifted forward and used to initialize the next planning cycle. The controller executes the leading portion of the optimized trajectory while the next MPC update is computed, and switches to the new trajectory once optimization completes, yielding receding-horizon control.
\subsection{Real-Time Digital Twin}
\label{sec:digital_twin}

The controller maintains a MuJoCo digital twin of the physical system, consisting of a Franka manipulator, an Allegro hand, and the manipulated object. Robot encoders and motion capture update the robot and object states in MuJoCo at approximately $100$~Hz, ensuring that each batch of candidate trajectories begins from the latest measured state. The resulting commands are transmitted to the low-level robot controllers at $333$~Hz, the maximum rate supported by the Allegro hand. Overall system architecture is visualized in Fig.~\ref{fig:overview}.

\subsection{Primitive-Residual Parameterization and Sampling}
\label{sec:primitive_residual}

Directly sampling every joint command independently is inefficient for dexterous manipulation because useful behaviors typically require coordinated motion across multiple fingers. We therefore parameterize the hand action using a low-dimensional set of motion primitives together with a joint-level residual.

From the trajectory data used for primitive extraction, we compute the mean hand posture
$\bar{\mathbf{q}}\in\mathbb{R}^{n_h}$ and extract a low-dimensional coordination basis
$\boldsymbol{\Phi}$ from the corresponding mean-centered joint trajectories.
\begin{equation}
\boldsymbol{\Phi}
=
\begin{bmatrix}
\boldsymbol{\phi}_1 & \cdots & \boldsymbol{\phi}_d
\end{bmatrix}
\in\mathbb{R}^{n_h\times d},
\end{equation}
where the columns of $\boldsymbol{\Phi}$ are $d$ coordinated motion primitives,
$n_h$ is the number of hand actuators, and $d\ll n_h$. The latent command at spline knot $k$ is
\begin{equation}
    \boldsymbol{\eta}_k
    =
    \begin{bmatrix}
        \mathbf{z}_k^\top &
        \delta\mathbf{q}_k^\top
    \end{bmatrix}^{\top},
    \label{eq:primitive_latent}
\end{equation}
where $\mathbf{z}_k\in\mathbb{R}^{d}$ contains the primitive mixing coefficients and
$\delta\mathbf{q}_k\in\mathbb{R}^{n_h}$ contains the joint-level residual. The corresponding hand command is reconstructed as
\begin{equation}
    \mathbf{u}_{h,k}
    =
    \bar{\mathbf{q}}
    +
    \sum_{j=1}^{d} z_{j,k}\boldsymbol{\phi}_j
    +
    \delta\mathbf{q}_k
    =
    \bar{\mathbf{q}}
    +
    \boldsymbol{\Phi}\mathbf{z}_k
    +
    \delta\mathbf{q}_k.
    \label{eq:primitive_residual}
\end{equation}

The coefficients in $\mathbf{z}_k$ are optimized jointly, allowing multiple primitives to contribute to the command at the same knot. Therefore, the controller fuses coordinated motion patterns rather than selecting a single primitive. The residual $\delta\mathbf{q}_k$ complements this fused primitive motion with joint-level adjustments. Thus, rather than reducing the total optimization dimension, the parameterization concentrates structured exploration along a low-dimensional, behavior-informed subspace while retaining smaller-scale full-dimensional exploration for local adaptation. Equivalently, under the Gaussian sampling used by CEM, it induces a behavior-informed low-rank-plus-diagonal sampling covariance in joint space, while the primitive-residual coordinates allow the two components to be controlled separately.


At each spline knot, CEM samples the primitive coefficients and joint-level residuals independently, with their sampling spreads determined by the corresponding components of $\boldsymbol{\sigma}$. After each CEM update, the standard deviations are recomputed from the elite samples. To prevent the sampling distribution from collapsing and to retain broad primitive-guided exploration together with smaller-scale joint-level refinement, we impose two standard-deviation floors, $\boldsymbol{\sigma}^{\mathrm{explore}}$ and $\boldsymbol{\sigma}^{\mathrm{refine}}$, with $\boldsymbol{\sigma}^{\mathrm{explore}}>\boldsymbol{\sigma}^{\mathrm{refine}}$. Half of the candidate trajectories are sampled using the exploration floor, while the remaining half use the refinement floor. These floors are defined separately for the primitive coefficients and joint-level residuals, as detailed in Sec.~\ref{sec:experimental_setup}.

\subsection{Primitive Extraction}
\label{sec:primitive_extraction}


The coordination basis is designed to capture patterns of joint coordination associated with a manipulation behavior, such as continuous rotation or grasping. It can be extracted from example trajectories exhibiting that behavior, including sensorized-glove demonstrations and trajectories generated by a policy trained in simulation. The data source used to construct each primitive is described in Sec.~\ref{sec:Results}.

The collected trajectories are temporally aligned and centered about their mean, $\bar{\mathbf{q}}$. Principal component analysis is then applied to the covariance of the centered joint trajectories to identify the dominant patterns of coordinated motion. Let $\lambda_j$ and $\mathbf{v}_j$ denote the eigenvalues and eigenvectors of the covariance matrix, ordered by decreasing eigenvalue. The coordination basis is formed as
\begin{equation}
    \boldsymbol{\Phi}
    =
    \begin{bmatrix}
        \mathbf{v}_1 & \cdots & \mathbf{v}_d
    \end{bmatrix},
    \qquad
    \frac{\sum_{j=1}^{d}\lambda_j}
         {\sum_j\lambda_j}
    \geq \gamma,
    \label{eq:primitive_extraction}
\end{equation}
where $\gamma$ is the selected explained-variance threshold. Once extracted, the coordination basis serves as an action-space prior, while the online MPC jointly optimizes its mixing coefficients and the joint-level residual according to the current task objective.

\subsection{Task-Constrained Rollout Evaluation}
\label{sec:constraints}

The task objective in~\eqref{eq:objective} provides a soft measure for ranking candidate trajectories, while task constraints specify conditions that must remain satisfied throughout a rollout. Such constraints can encode, for example, that the object remains within an admissible region. They are checked after every simulated transition rather than only at the end of the planning horizon.

If a candidate violates any enabled constraint, the virtual evaluation is immediately terminated and assigned a large failure cost, preventing it from entering the CEM elite set. The corresponding rollout worker can then begin evaluating a new sample without simulating the invalid candidate for the remainder of the horizon. This early-rejection mechanism increases the computational efficiency of rollout evaluation by directing the available resources toward trajectories that remain feasible and can contribute to the optimization.

\section{Experiments and Results}
\label{sec:Results}

\subsection{Experimental Setup}
\label{sec:experimental_setup}

\begin{figure}[t]
\centering
\includegraphics[width=\columnwidth]{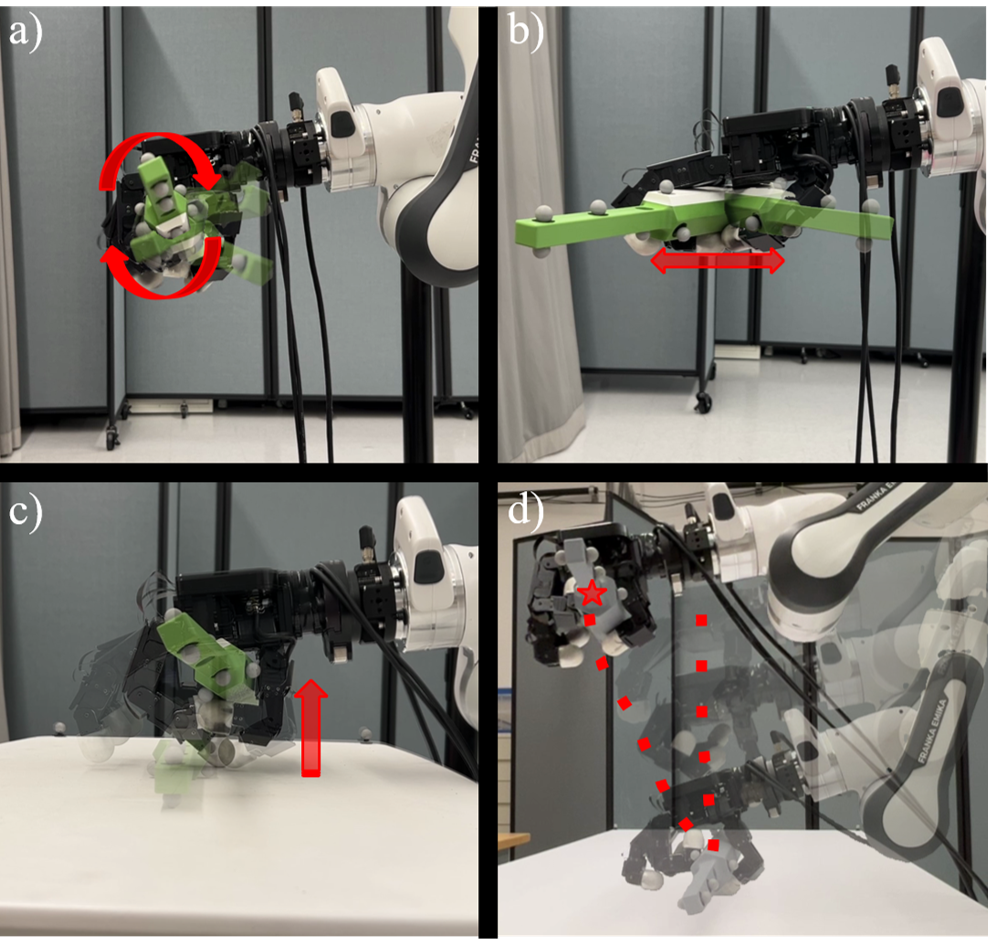}
\caption{Experimental setup. a) In-hand rotation, b) Reorientation, c) Grasping, d) Coordinated arm-hand manipulation.}
\label{fig:exp_setup}
\end{figure}

We evaluate the proposed primitive-informed sampling-based MPC framework on continuous in-hand rotation, object reorientation, grasping, and coordinated arm-hand reach-grasp-transport, as summarized in Fig.~\ref{fig:exp_setup}. Continuous rotation is used for the core ablations and robustness tests; reorientation isolates the effect of rollout constraints in goal tracking; grasping evaluates a human-derived primitive; and reach-grasp-transport extends the framework to coordinated arm-hand control.

The experiments examine: 1) the contributions of the manipulation primitive, joint-level residual, rollout constraints, and rollout budget; 2) whether a primitive extracted for one condition remains effective under changes in object geometry and model parameters; and 3) whether the framework can incorporate primitives from different data sources and extend from hand-only to coordinated arm-hand manipulation.

The MPC uses a $0.5$~s planning horizon represented by $K=6$ spline knots and $N_e=10$ elites per CEM update. The rollout budget $N$ is evaluated separately in Sec.~III-B2; based on this evaluation, $N=350$ is used for the subsequent hand-only experiments. For each joint-level residual dimension, the exploration and refinement standard-deviation floors are $\sigma_{\delta q}^{\mathrm{explore}}=1.0$~rad and $\sigma_{\delta q}^{\mathrm{refine}}=0.10$~rad. For primitive coefficient $j$, they are $\sigma_{z,j}^{\mathrm{explore}}=p_j^{\max}-p_j^{\min}$ and $\sigma_{z,j}^{\mathrm{refine}}=0.3\left(p_j^{\max}-p_j^{\min}\right)$, where $[p_j^{\min},p_j^{\max}]$ is the range of the $j$-th coefficient in the extraction data. Rollouts are evaluated in parallel on an AMD Ryzen Threadripper PRO 7995WX (192 threads), with each MPC update taking approximately 38~ms at $N=350$.

Statistical tests use trial-level samples: two-sided exact Wilcoxon rank-sum permutation tests for scalar and count outcomes, and two-sided Fisher exact tests for binary outcomes. Videos of the experiments are provided in the supplementary video and on the project webpage.

\subsection{In-Hand Rotation} 
\label{sec:rotation}

\subsubsection{Task Setup and Objective}
\label{sec:rotation_setup}



We evaluate continuous in-hand rotation of a $40$~mm across-flats (AF) hexagonal object about its principal axis. The MPC objective encourages rotation about this axis while maintaining the object's initial position and orientation about the remaining axes. The running cost is defined as
\begin{equation}
\ell_{\mathrm{rot}}(x_t)
=
w_p
\left\|
\mathbf{p}^o_t-\mathbf{p}^o_0
\right\|_2^2
+
w_{\parallel}
\left(
\psi_t-\psi^{\mathrm{des}}
\right)^2
+
w_{\perp}
\left\|
\mathbf{e}^{\perp}_t
\right\|_2^2 ,
\label{eq:rotation_cost}
\end{equation}
where $\mathbf{p}^o_t$ is the object position, $\mathbf{p}^o_0$ its initial position, $\psi_t$ its unwrapped principal-axis rotation, and $\mathbf{e}^{\perp}_t$ the orientation error about the remaining axes relative to the initial orientation. To maintain a persistent rotation command rather than reaching a fixed target angle, $\psi^{\mathrm{des}}$ is updated at each MPC replanning step to remain one full revolution ahead of the measured object orientation. A rotation attempt is successful if the object completes one full revolution without being dropped, and the rotation success ratio is the fraction of successful attempts within a trial.


In contrast to prior sampling-based MPC demonstrations of in-hand reorientation~\cite{li2024drop,hess2024sampling}, in which the palm faces upward and supports the object, the hand here holds the object without palm support (Fig.~\ref{fig:exp_setup}). A sampling update that fails to find a grasp-maintaining motion therefore results in a drop rather than a recoverable resting configuration, so coordinated finger motion must be identified at every replanning step.

The rotation primitive is extracted from trajectories generated by a Soft Actor-Critic (SAC) policy~\cite{haarnoja2018soft, kucuktabak2026realworld} trained in simulation using the same $40$~mm AF object, without domain randomization. We execute the trained policy in simulation and collect five trajectories totaling approximately one minute of motion. We then apply the primitive-extraction procedure described in Sec.~\ref{sec:primitive_extraction}. The minimum number of principal components explaining at least $90\%$ of the trajectory variance is retained, resulting in four primitives for the rotation task. Direct deployment of the simulation-trained policy fails to complete a full rotation on the physical system, likely because sim-to-real discrepancies in the contact dynamics disrupt the contact-rich finger-gaiting behavior. Nevertheless, the simulated trajectories retain useful kinematic finger coordination and provide an informative basis for primitive-guided sampling.

For this task, object-table contact is treated as a drop and therefore violates the rollout constraint.
If a virtual rollout violates this constraint, its evaluation is immediately terminated, as described in Sec.~\ref{sec:constraints}. 

\subsubsection{Effect of Rollout Budget}
\label{sec:rollout_budget}

\begin{figure}[t]
    \centering
    \includegraphics[width=\columnwidth]{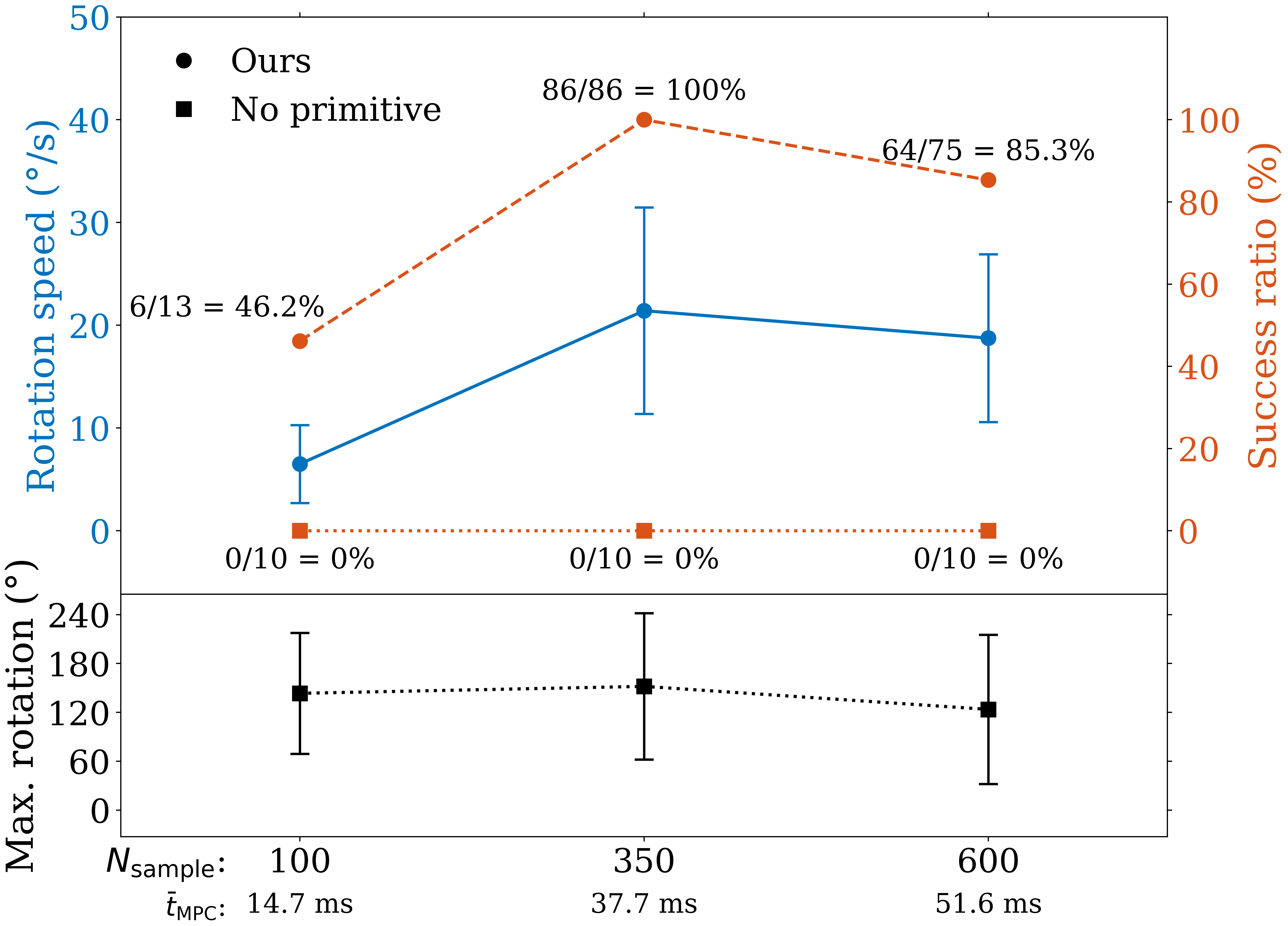}
    \caption{In-hand rotation performance versus sampling budget $N$. Top: rotation speed of successful rotations (left axis) and success ratio (right axis) for \emph{Ours} (circles) and \emph{No primitive} (squares). Bottom: maximum accumulated rotation per attempt for \emph{No primitive}. Mean MPC update time is shown below each $N$. Error bars denote $\pm$~SD.}
    \label{fig:sampling_sweep}
\end{figure}

We first evaluate how the rollout budget affects performance and whether increasing the number of samples can compensate for unstructured joint-space sampling. We vary $N\in\{100,350,600\}$ for both the complete method (\emph{Ours}) and a \emph{No primitive} variant, while keeping $N_e$, the planning horizon, and the number of spline knots fixed. In \emph{No primitive}, the controller samples directly in the joint-command space, as in prior sampling-based MPC for multi-fingered hands~\cite{li2024drop,hess2024sampling}. This comparison evaluates whether increasing the rollout budget is sufficient for direct joint-level sampling to discover the coordinated finger motions required for continuous rotation.
The complete method is evaluated over $10$ three-minute trials at each rollout budget, while \emph{No primitive} is evaluated over $10$ rotation attempts at each budget because it does not complete a full rotation at any tested budget (Fig.~\ref{fig:sampling_sweep}).

\claim{More samples do not replace the primitive.}
\emph{No primitive} achieves no successful rotations at any budget, and its maximum accumulated rotation remains approximately $120$-$150^\circ$ across the tested values of $N$ (Fig.~\ref{fig:sampling_sweep}). This plateau corresponds to the rotation achievable with the initial grasp. Continuing beyond it requires finger gaiting: the thumb and an opposing finger must release, reposition, and re-establish contact while the remaining fingers hold the object (see supplementary video). Independently sampled joint motions rarely produce this coordinated regrasp, so a six-fold increase in the number of sampled trajectories does not discover the coordination required to complete a full revolution.

In contrast, \emph{Ours} achieves a success ratio of $46.2\%$ ($6/13$) at $N=100$ and $100\%$ ($86/86$) at $N=350$. Increasing the budget further to $N=600$ raises the mean MPC update time from $37.7$~ms to $51.6$~ms and reduces the success ratio to $85.3\%$ ($64/75$). The longer update time reduces closed-loop reactivity and allows discrepancies between the simulated rollouts and the physical system to accumulate before replanning, offsetting the benefit of evaluating additional samples. We therefore use $N=350$ for the subsequent experiments.

\subsubsection{Component Ablations}
\label{sec:rotation_ablation}


We isolate the contributions of the manipulation primitive, joint-level residual, and rollout constraint by comparing \emph{Ours} against variants that remove each component.

For \emph{No primitive}, we additionally test $\sigma_{\delta q}^{\mathrm{explore}}\in\{0.5,1.0,1.5\}$~rad, with $\sigma_{\delta q}^{\mathrm{refine}}=0.1\,\sigma_{\delta q}^{\mathrm{explore}}$, to verify that its performance is not limited by a particular choice of joint-space sampling scale. This variance sweep is applied only to the \emph{No primitive} condition. In \emph{No residual}, we set $\delta q_k=0$ in~\eqref{eq:primitive_residual}, restricting all candidate motions to the primitive subspace. In \emph{No constraint}, the object-table contact constraint is disabled while the primitive-residual parameterization remains unchanged.

Conditions that achieve successful rotations (i.e., \emph{Ours} and \emph{No constraint}) are each evaluated over $10$ three-minute trials. Conditions with $0\%$ success (i.e., the three \emph{No primitive} sampling scales and \emph{No residual}) are each evaluated over $10$ rotation attempts. Across all evaluations, the initial object placement is varied by approximately $2$-$3$~cm.

\begin{figure}[t]
    \centering
    \includegraphics[width=\columnwidth]{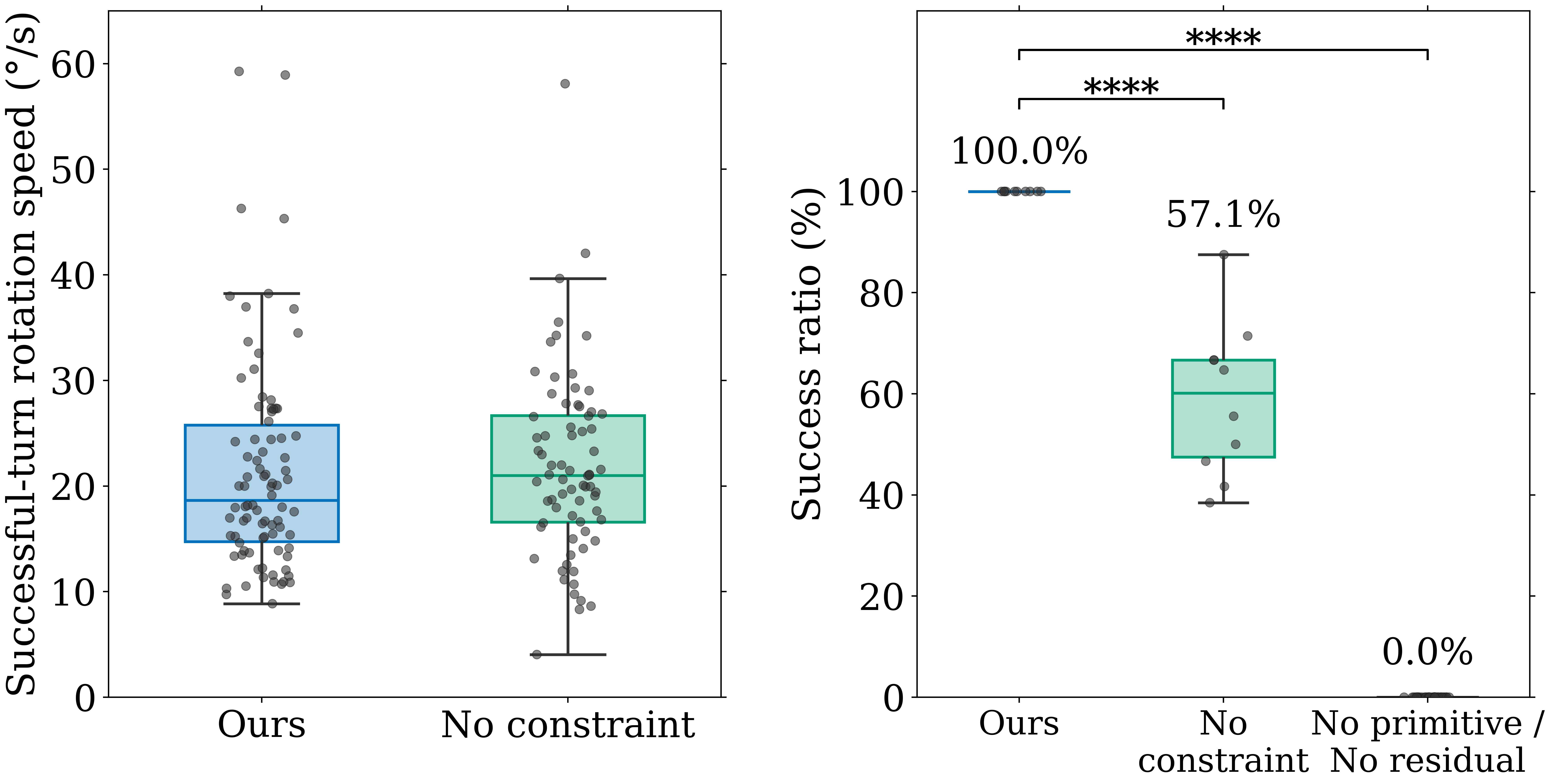}
    \caption{In-hand rotation ablations. Successful-turn rotation speeds are shown for \emph{Ours} and \emph{No constraint}, with each point representing one full rotation. Success ratios are shown for these two conditions and the pooled \emph{No primitive} and \emph{No residual} ablations. For the first two conditions, each success-ratio data point is computed over one three-minute trial.}
    \label{fig:rotation_ablation}
\end{figure}

\claim{Both primitive and residual are necessary.}
\emph{No residual} achieves no successful rotations across $10$ attempts, and none of the three \emph{No primitive} sampling scales completes a full rotation ($0/30$), confirming that the failure of joint-space sampling is not due to a particular choice of sampling scale. Because all four ablation settings produce the same outcome, their attempts are pooled for the aggregate success-rate comparison in Fig.~\ref{fig:rotation_ablation}. These conditions achieve a $0\%$ success rate, compared with $100\%$ ($86/86$ rotations across $10$ three-minute trials) for \emph{Ours}. The two ablations fail differently. As discussed in Sec.~\ref{sec:rollout_budget}, \emph{No primitive} fails to produce the coordinated regrasp required for finger gaiting and consequently loses the object, whereas \emph{No residual} retains the object in $9/10$ attempts and achieves a final rotation of $163.7^\circ \pm 123.8^\circ$, but eventually stalls. These results indicate that the primitive provides the coordinated motion structure needed to maintain manipulation through contact transitions, while the residual provides the local joint-level flexibility needed to adapt that motion as the hand-object configuration evolves.



\claim{Constraints improve reliability, not speed.}
Relative to the full method, disabling the rollout constraint reduces the success ratio to $57.1\%$ ($72/126$; $p=3.25\times10^{-5}$). However, the median speed among successful rotations remains similar: $20.42^\circ$/s without the constraint versus $18.78^\circ$/s with the complete method ($p=0.604$). The constraint therefore improves object retention without slowing the rotations that are successfully executed.

\subsubsection{Transfer Across Object Sizes}
\label{sec:object_size}


We evaluate the rotation primitive previously extracted using the $40$~mm AF object on $35$~mm and $45$~mm AF hexagonal objects without re-extraction. Each object size is evaluated over $10$ three-minute trials.
\begin{figure}[t]
    \centering
    \includegraphics[width=1\columnwidth]{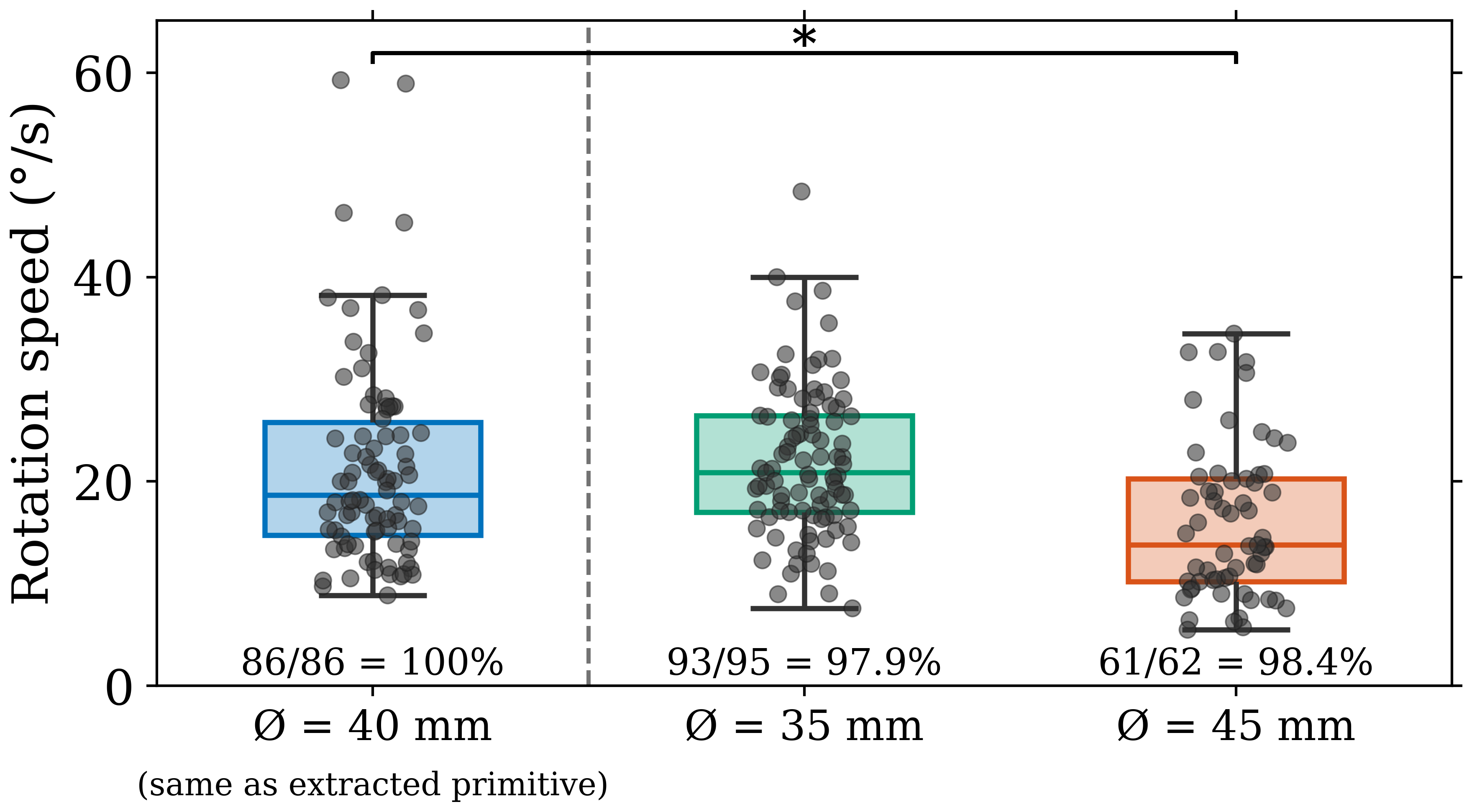}
    \caption{In-hand rotation performance across object sizes using the primitive extracted from the $40$~mm AF object. Successful-turn rotation speeds and rotation success ratios are shown.}
    \label{fig:object_size}
\end{figure}

\claim{The primitive transfers across object sizes.}
The rotation success ratio remains high for both object sizes different from that used for primitive extraction, with $93/95$ successful rotations ($97.9\%$) for the $35$~mm object and $61/62$ ($98.4\%$) for the $45$~mm object, compared with $86/86$ ($100\%$) for the $40$~mm object (Fig.~\ref{fig:object_size}). The $35$~mm and $45$~mm objects are dropped twice and once, respectively, compared with no drops for the $40$~mm object; these differences are not statistically significant ($p=0.34$). Rotation speed is unchanged for the $35$~mm object ($p=1.00$), while the median speed decreases from $18.78^\circ$/s for the $40$~mm object to $13.17^\circ$/s for the $45$~mm object ($p=0.046$). These results show that a primitive extracted using one object size remains informative across the tested size variations, while the joint-level residual allows the MPC to adapt the primitive to the corresponding hand-object configurations.

\subsubsection{Robustness to Model Mismatch}
\label{sec:model_mismatch}

We evaluate sensitivity to model error by independently perturbing either the object friction coefficient or mass in the MPC digital twin by $\pm50\%$ relative to its nominal value, while leaving the physical task unchanged. The nominal model and each of the four perturbed models are evaluated over $10$ three-minute trials.
\begin{figure}[t]
    \centering
    \includegraphics[width=\columnwidth]{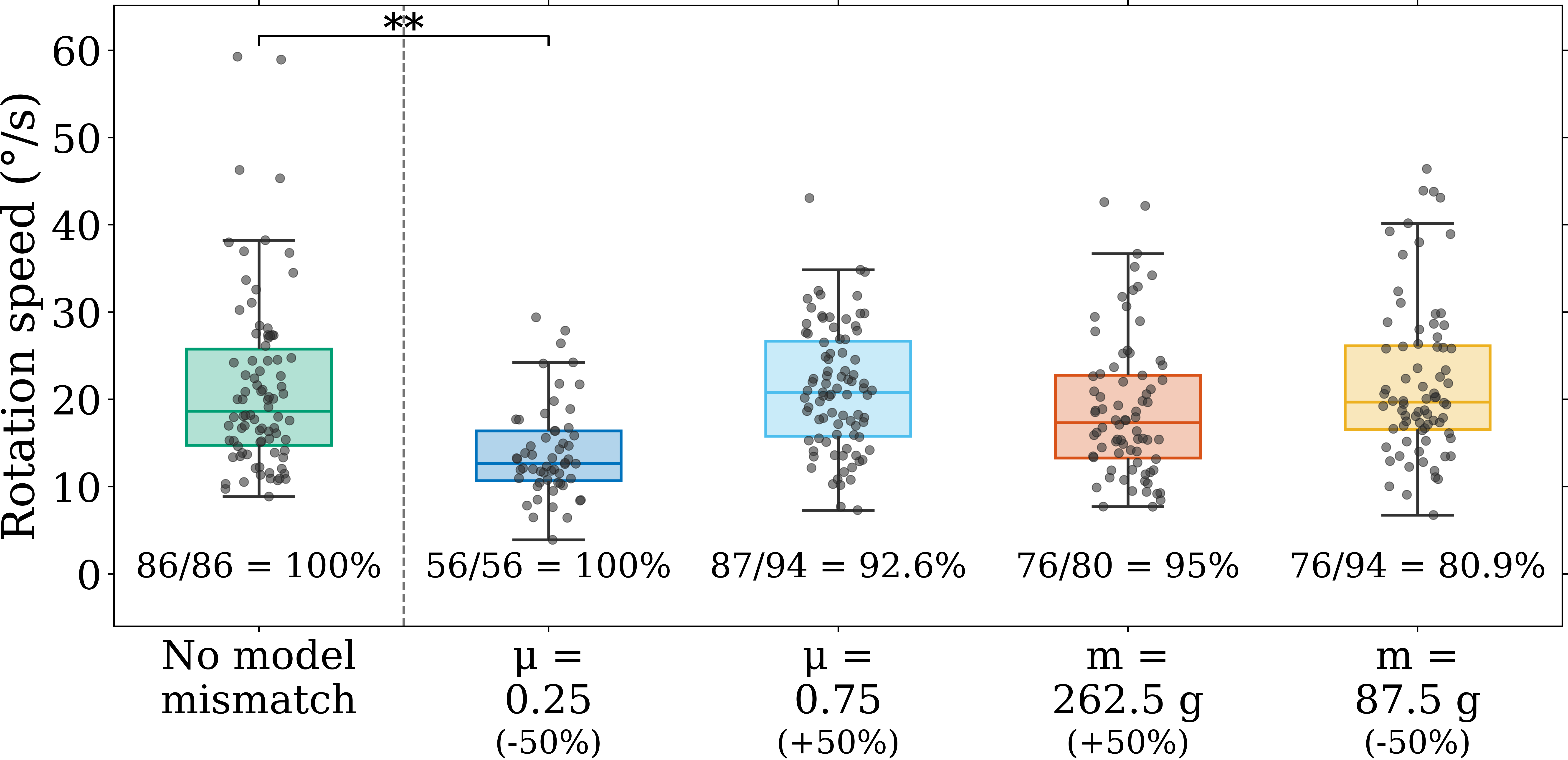}
    \caption{In-hand rotation performance under model mismatch. The friction coefficient and object mass represented in the MPC digital twin are varied by $\pm50\%$ relative to their nominal values. Successful-turn rotation speed and the number of object drops are shown for each condition.}
    \label{fig:model_mismatch}
\end{figure}

\claim{Rotation persists under 50\% model mismatch.}
The method produces repeated in-hand rotations under all four mismatch conditions, with rotation success ratios between $80.9\%$ and $100\%$ (Fig.~\ref{fig:model_mismatch}). Underestimating friction by $50\%$ reduces the median rotation speed from $18.78^\circ$/s under the nominal model to $13.03^\circ$/s ($p=0.002$), while the remaining perturbations do not significantly affect rotation speed. Object retention is most affected when the model underestimates the object mass by $50\%$, producing $18$ drops over $30$ minutes, significantly more than under the nominal model ($p=9.53\times10^{-4}$); drop counts under the other perturbations do not differ significantly from nominal. These results show that the receding-horizon controller remains effective under substantial errors in the modeled object parameters, although sufficiently large modeling errors degrade rotation speed or object retention.

\subsection{Grasping}
\label{sec:grasping}

For the grasping task (Fig.~\ref{fig:exp_setup}(c)), the hand starts from an open configuration above the same $40$~mm AF hexagonal object used in the rotation experiments, with the object resting on the table. The MPC objective encourages the hand to lift the object while maintaining its initial orientation. The running cost is defined as
\begin{equation}
\ell_{\mathrm{grasp}}(x_t)
=
10
\left(
z_t^o-z^{o,\mathrm{des}}
\right)^2
+
\left\|
\mathbf{e}^o_t
\right\|_2^2 ,
\label{eq:grasp_cost}
\end{equation}
where $z_t^o$ is the vertical object position, $z^{o,\mathrm{des}}$ corresponds to a position $10$~cm above the hand, and $\mathbf{e}^o_t$ is the quaternion orientation error relative to the initial object orientation.

The primitive basis is extracted from human hand-motion data from the NinaPro database~\cite{atzori2012ninapro}. Specifically, we use trajectories from $25$ right-handed subjects performing $20$ different grasp motions. Human finger flexion angles are mapped directly to the corresponding Allegro flexion joints of the index, middle, ring finger, and thumb, without kinematic retargeting; the little finger is discarded, and the abduction-adduction joints, which lack a direct counterpart in the glove data, are excluded from the basis and driven only by the joint-level residual.

The resulting trajectories capture coordinated hand configurations across a diverse set of grasps and are processed using the same PCA-based primitive extraction procedure described in Sec.~\ref{sec:primitive_extraction}. Applying the $90\%$ explained-variance threshold results in five primitives for the grasping task. To evaluate the contribution of this coordination prior, we conduct $10$ trials per condition, comparing \emph{Ours} with \emph{No primitive}, defined as in Sec.~\ref{sec:rotation_ablation}. Each trial starts from the same open-hand configuration, and a grasp is considered successful if the object is lifted at least $5$~cm from the table.

\claim{A human-motion primitive enables grasping.}
The complete method succeeds in $9$ of $10$ trials, whereas \emph{No primitive} fails in all $10$ trials ($p=1.19\times10^{-4}$). Without the primitive, sampled motions typically push or tilt the object rather than producing the coordinated finger closure needed to establish a grasp. 
The grasping primitive differs from the rotation primitive in both source and behavior: it is extracted from human hand-motion data rather than a simulation-trained policy and captures grasping coordination rather than finger gaiting. Together with the rotation experiments, these results show that the framework can exploit coordination priors from different data sources for their respective manipulation behaviors.

\subsection{Object Reorientation}
\label{sec:reorientation}

We further evaluate the rollout constraint on a repetitive heading-tracking task (Fig.~\ref{fig:exp_setup}(b)) in which the desired object heading traverses a $60^\circ$ range over $10$~s, is held for $3$~s, and then reverses direction. The running cost is defined as
\begin{equation}
\ell_{\mathrm{reorient}}(x_t)
=
5
\left\|
\mathbf{p}^o_t-\mathbf{p}^o_0
\right\|_2^2
+
1.5
\left\|
\mathbf{e}^o_t
\right\|_2^2 ,
\label{eq:reorientation_cost}
\end{equation}
where $\mathbf{p}^o_t$ is the object position, $\mathbf{p}^o_0$ its initial position, and $\mathbf{e}^o_t$ the quaternion orientation error relative to the commanded object pose.

Unlike continuous rotation, this task does not require finger gaiting, and comparatively small joint motions can achieve the commanded heading changes. Direct joint-space sampling is therefore sufficient for this task, allowing the effect of the rollout constraint to be evaluated independently of primitive-informed sampling. As in the rotation experiments, a rollout is rejected if the object contacts the table.

We conduct $10$ trials with the rollout constraint enabled and $10$ with it disabled. We compare full-sequence completion and trial duration before an object drop, and evaluate tracking performance using the heading root-mean-square error (RMSE) during transitions and the mean reaching error during the $3$~s hold periods.

\begin{figure}[t]
    \centering
    \includegraphics[width=\columnwidth]{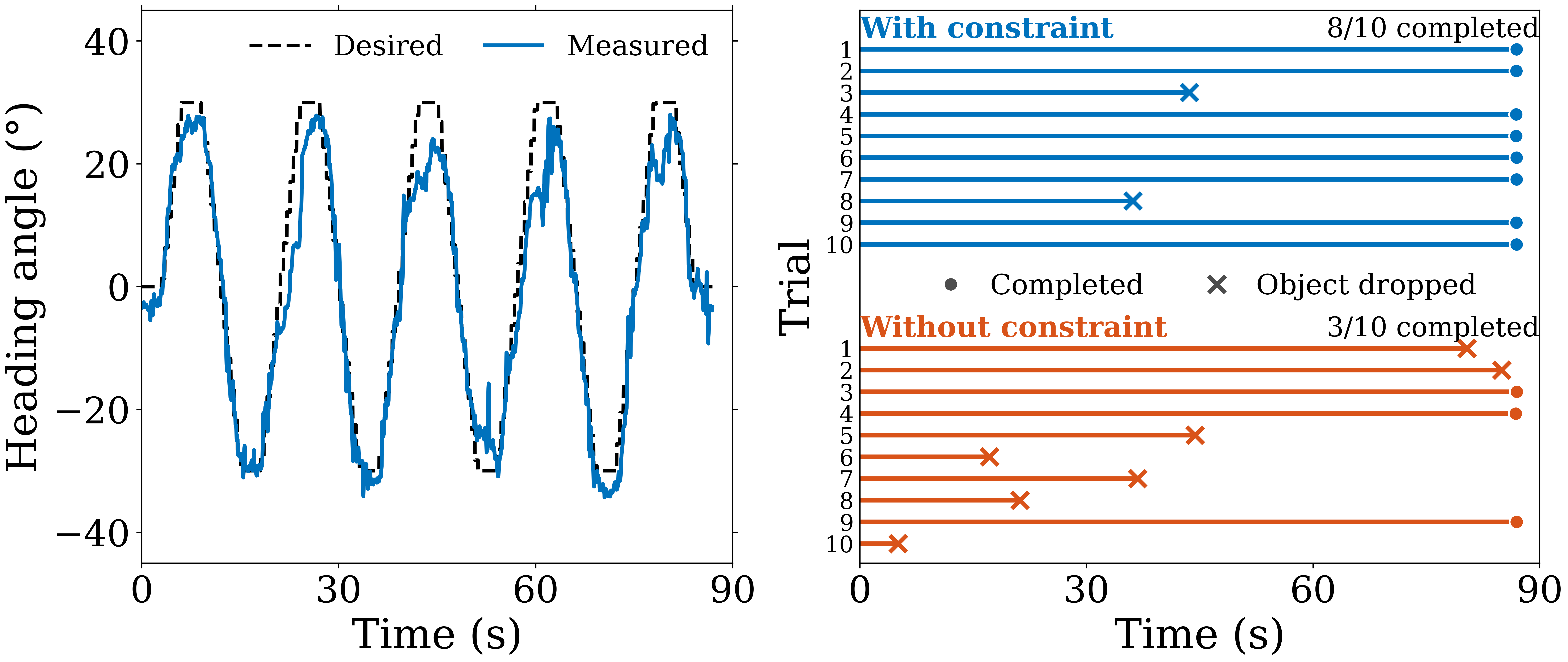}
    \caption{Object reorientation with and without rollout constraints. Left: desired and measured object heading for a representative trial with the rollout constraint enabled. Right: active duration and outcome of each trial. Circles indicate completion of the full reorientation sequence, while crosses indicate termination due to an object drop.}
    \label{fig:orientation}
\end{figure}

\claim{Constraints decrease drops.}
The controller tracks the cyclic heading command while repeatedly reversing the direction of object rotation (Fig.~\ref{fig:orientation}). With the rollout constraint enabled, $8$ of $10$ trials complete the full sequence, compared with $3$ of $10$ without it. Every non-completed trial ends in an object drop, and these drops occur earlier when the constraint is disabled (Fig.~\ref{fig:orientation}). Tracking accuracy while the object remains in the hand is similar between the two conditions: the median heading RMSE is $7.87^\circ$ with the constraint and $8.26^\circ$ without ($p=0.993$), and the median mean reaching error is $1.88^\circ$ and $1.72^\circ$, respectively ($p=0.993$). As in the in-hand rotation ablation, the rollout constraint therefore affects object retention rather than the accuracy of the reorientation motions that are executed.

\subsection{Coordinated Arm-Hand Manipulation}
\label{sec}

We further evaluate the framework on a coordinated arm-hand manipulation task (Fig.~\ref{fig:exp_setup}(d)). The object is assigned a desired pose in the world frame, while the robot starts from a configuration in which the hand is away from the object. The task therefore requires coordinated motion of the Franka arm and Allegro hand to reach the object, establish a grasp, and transport it toward the desired pose.

We use the same grasping primitive extracted from the NinaPro human hand-motion data in Sec.~\ref{sec:grasping}. For this experiment, the primitive-residual action parameterization in~\eqref{eq:primitive_latent} is extended by augmenting each latent spline knot with a desired Cartesian pose for the Franka end effector,
\begin{equation}
    \boldsymbol{\eta}_k
    =
    \begin{bmatrix}
        \mathbf{z}_k^\top &
        \delta\mathbf{q}_k^\top &
        \mathbf{p}_{\mathrm{EE},k}^{\mathrm{des}\top} &
        \mathbf{q}_{\mathrm{EE},k}^{\mathrm{des}\top}
    \end{bmatrix}^{\top},
    \label{eq:arm_hand_latent}
\end{equation}
where $\mathbf{p}_{\mathrm{EE},k}^{\mathrm{des}}\in\mathbb{R}^{3}$ and $\mathbf{q}_{\mathrm{EE},k}^{\mathrm{des}}\in S^{3}\subset\mathbb{R}^{4}$ denote the desired end-effector position and quaternion orientation at spline knot $k$, respectively. During CEM sampling, end-effector orientations are perturbed in axis-angle form and interpolated between spline knots using shortest-path spherical linear interpolation.

A simple task objective penalizes the error between the object and its desired pose in the world frame together with the error between the object pose relative to the wrist and a desired object-wrist configuration. Let $\mathbf{e}^{W}_{o,t}$ denote the object-pose error relative to the desired world-frame pose and $\mathbf{e}^{\mathrm{EE}}_{o,t}$ denote the error relative to the desired object pose in the wrist frame. The running cost is
\begin{equation}
\ell_{\mathrm{arm\text{-}hand}}(x_t)
=
w_W
\left\|
\mathbf{e}^{W}_{o,t}
\right\|_2^2
+
w_{\mathrm{EE}}
\left\|
\mathbf{e}^{\mathrm{EE}}_{o,t}
\right\|_2^2 .
\label{eq:arm_hand_cost}
\end{equation}
The world-frame term drives the object toward its target pose, while the wrist-relative term encourages a suitable hand-object configuration during manipulation.

Due to the increase in action-space dimension introduced by the Cartesian arm variables, we increase the number of candidate trajectories evaluated at each MPC update to $N=500$. The planning horizon and number of spline knots remain unchanged.

A trial is considered successful if the object reaches within $3$~cm of the desired position. The controller establishes a grasp in $9$ of $10$ trials and successfully transports the object to the target in $8$ of $10$ trials. In one trial, the object is successfully grasped and lifted but is dropped during transport. 
Although task completion requires the robot to reach toward the object, establish a grasp, and transport the grasped object to the desired position in sequence, these stages are not explicitly programmed or specified to the controller. Instead, the sequence of reaching, grasping, and transport emerges from online optimization of the task objective. The supplementary video shows representative executions, including one in which the robot first rotates the object in a non-prehensile manner before grasping and transporting it.
\section{Conclusion}
\label{sec:conclusion}

We present a primitive-informed sampling-based MPC framework for multi-fingered dexterous manipulation. The framework structures online trajectory sampling using low-dimensional manipulation primitives while jointly optimizing joint-level residuals and rejecting infeasible trajectories through task-related rollout constraints. Physical experiments show that the primitive and residual play complementary roles: the primitive provides the coordinated finger motion needed to efficiently explore difficult contact transitions, while the residual allows the MPC to adapt this coordination to the current hand-object configuration. Increasing the rollout budget alone does not recover the coordination provided by the primitive, and rollout constraints substantially improve object retention without degrading the quality of successful motions.

The same framework supports continuous in-hand rotation, grasping, object reorientation, and coordinated arm-hand manipulation. The rotation primitive remains effective across the tested object-size variations and under substantial model mismatch, while the grasping experiments show that useful coordination priors can also be extracted from human hand-motion data. These results suggest that low-dimensional motion structure can serve as an effective sampling prior without restricting manipulation to a fixed primitive or policy, leaving the task objective and local adaptation to online MPC. Future work will investigate broader object geometries and manipulation tasks, as well as methods for automatically constructing, adapting, and combining manipulation primitives.

\addtolength{\textheight}{-0cm}   

\bibliographystyle{bibliography/myIEEEtran} 
\bibliography{bibliography/references}

\end{document}